\documentclass[letterpaper, 10 pt, conference]{ieeetran}
\IEEEoverridecommandlockouts
\usepackage{times}

\usepackage{multicol}
\usepackage[bookmarks=true]{hyperref}

\usepackage{hyperref}
\usepackage{graphicx}		
\usepackage{wrapfig}
\usepackage[format=plain,font=footnotesize,labelfont=bf,labelsep=period]{caption}
\usepackage[export]{adjustbox}
\usepackage{bm}
\usepackage{subcaption}
\usepackage{balance}
\usepackage{dsfont}

\usepackage{amsmath}
\usepackage{mathrsfs}
\usepackage{amssymb}
\usepackage{mathtools}
\usepackage{relsize}
\usepackage{bbm}
\usepackage{cite}
\usepackage{algorithm}
\usepackage{algpseudocode}

\usepackage{pdfsync}
\usepackage[normalem]{ulem}
\usepackage{paralist}	
\usepackage[space]{grffile} 

\usepackage[usenames,dvipsnames]{xcolor}

\usepackage{amsthm}
\usepackage{centernot}

\newtheorem{theorem}{Theorem}
\newtheorem*{theorem*}{Theorem}

\newtheorem*{lemma*}{Lemma}
\newtheorem{definition}{Definition}
\newtheorem{proposition}{Proposition}
\newtheorem*{proposition*}{Proposition}

\newcommand{\newsec}[1]{\vspace{0.1cm} \noindent \textbf{#1} }

\newcommand{\R}{\mathbb{R}}

\newcommand{\C}{\mathcal{C}}

\definecolor{darkblue}{RGB}{0,0,102}
\definecolor{lightblue}{RGB}{77,77,148}

\definecolor{gold}{RGB}{234, 170, 0}
\definecolor{metallic_gold}{RGB}{139, 111, 78}

\newcommand{\mb}[1]{\mathbf{ #1 }}
\newcommand{\bs}[1]{\boldsymbol{ #1 }}

\DeclareMathOperator*{\argmin}{argmin}

\newcommand{\lmat }{\begin{bmatrix}}
\newcommand{\rmat}{\end{bmatrix}}

\newcommand{\E}{\mathbb{E}} 
\usepackage{dsfont}
 
\usepackage{svg}
\usepackage{caption}

\makeatletter
\long\def\@makefntext#1{%
  \noindent
  #1}
\makeatother

\begin{document}

\title{SHIELD: \underline{S}afety on \underline{H}umanoids via  CBFs \underline{I}n \underline{E}xpectation on  \\ \underline{L}earned \underline{D}ynamics}

    \author{Lizhi Yang$^{*1}$, Blake Werner$^{*1}$, Ryan K. Cosner$^1$, \\ David Fridovich-Keil$^2$, Preston Culbertson$^3$, and Aaron D. Ames$^{1}$ \thanks{$^1$Mechanical and Civil Engineering, California Institute of Technology \newline $^2$Aerospace Engineering and Engineering Mechanics, UT Austin  \newline $^3$Computer Science, Cornell University \newline $^*$ represents equal contribution
    \newline This research is supported in part by the Technology Innovation Institute (TII), BP p.l.c., and by Dow via project \#227027AW.}
}

\maketitle

\begin{abstract}
Robot learning has produced remarkably effective ``black-box'' controllers for complex tasks such as dynamic locomotion on humanoids. 
Yet ensuring dynamic safety, i.e., constraint satisfaction, remains challenging for such policies. 
Reinforcement learning (RL) embeds constraints heuristically through reward engineering, and adding or modifying constraints requires retraining. 
Model-based approaches, like control barrier functions (CBFs), enable runtime constraint specification with formal guarantees but require accurate dynamics models.
This paper presents SHIELD, a layered safety framework that bridges this gap by: (1) training a generative, stochastic dynamics residual model using real-world data from hardware rollouts of the nominal controller, capturing system behavior and uncertainties; and (2) adding a safety layer on top of the nominal (learned locomotion) controller that leverages this model via a stochastic discrete-time CBF formulation enforcing safety constraints in probability.  
The result is a minimally-invasive safety layer that can be added to the existing autonomy stack to give probabilistic guarantees of safety that balance risk and performance.  
In hardware experiments on an Unitree G1 humanoid, SHIELD enables safe navigation (obstacle avoidance) through varied indoor and outdoor environments using a nominal (unknown) RL controller and onboard perception. 

\end{abstract}

\IEEEpeerreviewmaketitle

\section{Introduction}
\begin{figure}[h]
\vspace{20pt}
    \centering
    \includegraphics[width=\linewidth]{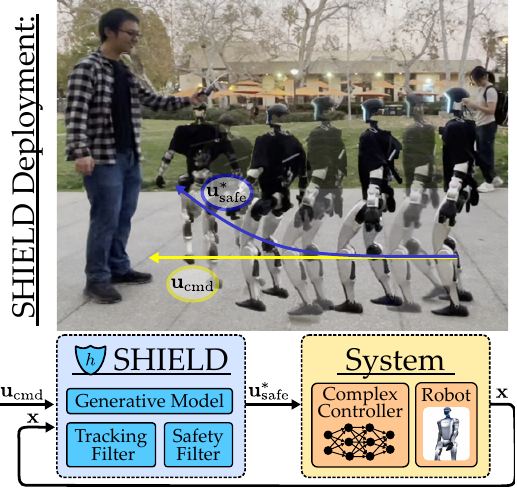}
    \caption{A humanoid robot implementing the SHIELD architecture autonomously avoids collision with a human using onboard sensing. SHIELD combines a performant underlying controller (e.g., an RL-trained locomotion policy) with a safety layer, which modulates high-level reference signals through a generative model of tracking error trained using real-world trajectory data. This architecture allows safety constraints (like collision avoidance) to be specified and enforced at runtime, with rigorous probabilistic guarantees, even on high-dimensional systems like humanoid robots with complex or ``black-box'' control policies.}
    \label{fig:hero}
    \vspace{-3em}
\end{figure}

As learning-based controllers achieve remarkable success in complex robotic tasks such as legged locomotion \cite{hwangbo2019Learning, lee2020learning, feng2023genloco,liao2024berkeley, miki2022learning, radosavovic2024real,zakka2025mujoco,zhuang2024humanoid}, they bring with them a fundamental tension: the black-box, data-driven nature, which enables their robust performance, simultaneously obscures our ability to provide formal safety guarantees or modify their constraints without expensive retraining. 
As more roboticists begin to field robust controllers trained using strategies like reinforcement learning (RL), developing ways to flexibly and adaptively constrain their behavior online to ensure safety (e.g., to avoid colliding with humans in their workspace) remains an open problem.  Solving this problem is especially critical for humanoid robots, which by their very nature are designed to interact with humans in everyday environments. 

\newsec{Background.}  Several methods have emerged in recent years to enable the safe deployment of learning-based controllers. For example, conformal prediction provides a powerful framework for developing risk-aware controllers with quantile-based robustification \cite{lindemann2023safe,akella2024sample}. However, this approach often becomes computationally intractable as the number of samples increases.
Alternatively, ``backup''-style approaches employ a dual-controller strategy: a ``performant'' controller during normal operation and a separate safe controller that engages in high-risk scenarios, e.g., \cite{he2024agile} proposed using a learned safe controller, which inherits the same unpredictability as the ``performant'' controller when operating outside its training distribution. 
The safe controller can alternatively be designed using optimal control techniques like backward reachability via the Hamilton-Jacobi-Bellman (HJB) equations \cite{mitchell2005time}, but these methods rely heavily on accurate dynamics models and often prove computationally prohibitive \cite{bansal_hj_2017} for complex systems such as bipedal robots.
The above methods can be broadly framed under the notion of data-driven \emph{safety filters} \cite{wabersich2023data}, i.e., methods that modulate nominal signals to ensure safety in a data-driven context.  

\begin{figure*}[t!]
    \centering
    \includegraphics[width=\linewidth]{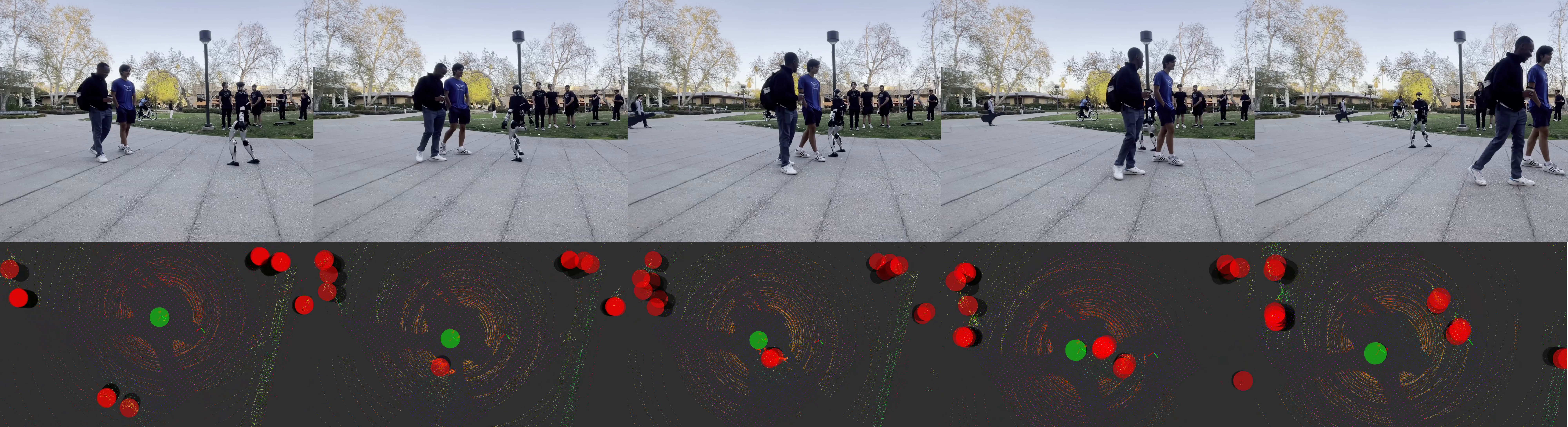}
    \caption{SHIELD enables real-world pedestrian avoidance with a humanoid robot, using a ``general-purpose'' RL policy. \emph{Top:} Our robot safely walks among pedestrians using SHIELD's stochastic safety framework. \emph{Bottom:} The robot relies solely on onboard perception to detect and avoid obstacles. Experimental video of this experiment can be found at: \href{https://vimeo.com/1061676063}{https://vimeo.com/1061676063}. }
    \label{fig:pedstrian}
    \vspace{-1.75em}
\end{figure*}

The concept of a safety filter originated with control barrier functions (CBFs) \cite{ames2017control,ames2019control}.  
This method takes a nominal controller (potentially learning-based) and filters it via the CBF condition to ensure safety framed as forward set invariance. 
This approach has proven effective on a wide-variety of robotic systems, including quadrupedal and bipedal robots \cite{grandia2021multi,csomay2021episodic}.  
Yet this approach assumes an accurate model of the system dynamics and environment---this is not available for complex humanoids operating in unstructured environments.  
To address this, recent work has leveraged reduced-order models in the synthesis of CBF-based safety filters \cite{molnar2021model,cohen2024reducedorder}, but this requires the underlying assumption of accurate tracking of reference signals. 

\newsec{Contributions.} This paper introduces SHIELD, a novel paradigm for guaranteeing safety in robotic systems that bridges the gap between data-driven and model-based safety methods. SHIELD is specifically designed for systems with complex, robust, but ultimately stochastic low-level controllers, such as RL policies used by humanoid robots for locomotion. Unlike traditional safety filters, SHIELD functions as a safety layer that sits ``above'' the nominal learning-based controller in the autonomy stack (cf. Fig. \ref{fig:hero}), modulating the reference signal rather than directly filtering control outputs. SHIELD is constructed through a three-step process:
\begin{enumerate}
    \item[\emph{Step 1: Constraint specification.}] The user specifies a safety constraint on a subset of the robot states (e.g. the pose of the robot torso) mathematically, with positive values corresponding to constraint satisfaction. The low-level policy does not need to be trained to satisfy this constraint but can instead be designed to track general reference commands provided to the reduced-order model (as is typical for RL \cite{lee2020learning, miki2022learning}).
    \item[\emph{Step 2: Dynamics residual learning.}] The user collects real-world data of the low-level policy being excuted and trains a conditional variational autoencoder (CVAE) to model the difference between the desired motion of the reduced-order model, and closed-loop system's real-world tracking of these commands.  The result is improved reference signal tracking performance. 
    \item[\emph{Step 3: Safety-aware reference generation.}] The learned residual distribution from the CVAE is used to compute ``minimally-invasive'' modifications to the reference command that closely track the desired motion of the reduced-order model while satisfying a stochastic discrete-time control barrier function (S-DTCBF) \cite{clark2019control,cosner2023robust} constraint. 
    The result is a formal guarantee of safety in probability: the probability that the system state leaves a specified safe set in a finite specified horizon.
\end{enumerate}
Crucially, in contrast to prior work \cite{cosner2024generative}, SHIELD uses the CVAE to both improve the qualitative performance (by achieving better tracking) and thereby, through the layered implementation, enforce user-specified safety constraints. The result is guarantees of safety in probability. 
We also provide a computationally tractable formulation of the S-DTCBF constraint for obstacle avoidance that is amenable to online computation on embedded robot hardware.

We validate our theoretical framework by implementing our approach on an Unitree G1 Humanoid robot and conducting comprehensive experiments. We first model the robot as a planar single integrator system, train an RL policy to track reference linear and yaw rates, and define obstacle avoidance constraints for these states using onboard perception (Step 1). We then train a CVAE to model these disturbances using obstacle-free locomotion data (Step 2). Online, we deploy the controller using a stochastic DTCBF with the generative dynamics residual (Step 3), which modulates the inputs to the RL walking policy. In controlled experiments, SHIELD consistently outperforms traditional DTCBF methods, as the robot tracks velocity commands while avoiding obstacles using only onboard perception. Finally, to demonstrate real-world applicability, we successfully deploy our system in unstructured outdoor environments (see Fig. \ref{fig:pedstrian}) where the robot navigates safely around humans. 
\section{background}

In this work we consider robots that can be modeled as discrete time dynamical systems of the form: 
\begin{align}
    \mb{s}_{k+1} = \bs{\Phi}(\mb{s}_k, \mb{a}_k). 
\end{align}
where $\mb{s}_k\in \R^{n_s}$ is the state of the system and $\mb{a} \in \R^{n_a}$ is the system input. This may be the high-dimensional representation of the system where $\mb{s}$ includes global pose, joint angles, joint angular velocities, etc. and $\mb{a}$ may be joint torques, voltages, etc.  For this complex system, we assume that we have some controller $\bs{\pi}: \R^{n_s} \times \R^{n_u} \to \R^{n_a}$ that takes the current system state $\mb{s}$ and user commands $\mb{u}$ to produce full-order system inputs $\mb{a}$. 
Using this controller yields: 
\begin{align}
    \mb{s}_{k+1} = \bs{\Phi}(\mb{s}_k, \bs{\pi}(\mb{s}_k,\mb{u}_k)  ) . 
\end{align}

For navigation purposes, we consider a reduced-order representation of the system $\mb{x} \in \R^{n_x}$ where $n_x < n_s$ and $\mb{x} = \mb{p}(\mb{s})$ for some projection $\mb{p}: \R^{n_s} \to \R^{n_x}$ that projects the full-order state $\mb{s}$ onto the reduced-order state $\mb{x}$. Here $\mb{x}$ may be the outputs of the system that are considered in safety and navigation, such as its center of mass position. Similarly, we consider the reduced-order inputs to the system. 

We can then represent the discrete-time dynamics of this reduced-order model of the system as: 
\begin{align}
\label{rom_system}
    \mb{x}_{k+1} & = \mb{p}(\bs{\Phi}(\mb{s}_k, \bs{\pi}(\mb{s}_k,\mb{u}_k) )) \\
    & \approx \mb{F}(\mb{x}_k) + \mb{G}(\mb{x}_k)\mb{u}_k + \mb{d}_k \label{eq:reduced_order_sys}
\end{align} 
where $\mb{F}(\mb{x}_k) + \mb{G}(\mb{x}_k)\mb{u}_k$ represents a simplified model of the system and $\mb{d}$ is the difference between the full-order model and this reduced order model, also called the dynamics residual. To capture the complexities of the full-order dynamics $\bs{\Phi}$ and the controller $\bs{\pi}$, we consider $\mb{d}_k$ to be a random disturbance sampled from a distribution $\mathcal{D}(\mb{s}_{k:0}, \mb{a}_{k:0})$ that is dependent on the history of full states and from time $0$ through $k$, denoted as $\mb{s}_{k:0}$  and $\mb{a}_{k:0}$ respectively. 

\newsec{Safety in Probability.} To consider the safety of this stochastic RL-guided system, we consider a system to be safe as long as its state is in a user defined \textit{safe set} $\mathcal{C}\subset \R^{n_x}$. 
Due to the stochasticity of our system, it may not be possible to guarantee with complete certainty that our system will remain in $\mathcal{C}$ for all time \cite[Sect. IV]{culbertson2023issp}. Instead, we look to bound finite-time safety probability as a metric for system safety, as is common in the stochastic safety literature \cite{santoyo_barrier_2021, steinhardt_stochastic_2012, cosner2023robust}.\footnote{Given the discrete nature of our problem formulation, we focus exclusively on safety at sample times as in \cite{cosner2023robust}. We refer to \cite{breeden_sampledDataCbfs_2022} for an analysis of inter-sample safety.}

\begin{definition}[$K$-step Exit Probability] 
    For any $K\in \mathbb{N}_1$ and initial condition $\mb{x}_0 \in \R^n$, the $K$-step exit probability of the set $\C$ for a feedback controller $\mb{u}_k = \mb{K}(\mb{x}_k)$  applied to the system \eqref{rom_system} is: 
    \begin{align}\label{eq:k_step_safety}
        P_{\mb{u}}(K, \mb{x}_0) = \mathbb{P} \left\{ \mb{x}_k \notin \mathcal{C} \textup{ for some }  k \leq K \right\} 
    \end{align}
\end{definition}

\newsec{Stochastic Safety Filters.} To enforce a bound on this $K$-step exit probability, we first define the safe set $\mathcal{C}$ as the 0-superlevel set of some function $h: \R^{n_x} \to \R$: 
\begin{equation}
    \mathcal{C} = \{\mb{x} \in \mathbb{R}^{n_x} ~|~ h(\mb{x}) \geq 0\}. 
\end{equation}
Using this definition of safety, the field nominally considers the classical discrete-time control barrier function (DTCBF) inequality to enforce safety: for $\alpha\in(0, 1)$\cite{agrawal_dtcbf_2017}:
\begin{align}
    h(\mb{x}_{k+1})\geq \alpha h(\mb{x}_k). \label{eq:nom_dtcbf_ineq} \tag{DTCBF}
\end{align}
However, given the stochastic nature of our system, we instead consider the following stochastic discrete-time control barrier function (S-DTCBF) inequality to enforce safety guarantees on our system: 
\begin{align}
    \E[ ~h(\mb{x}_{k+1}) ~|~ \mathscr{F}_{k} ~] \geq \alpha h(\mb{x}_k) \label{eq:dtcbf_ineq} \tag{S-DTCBF}
\end{align}
where $\mathscr{F}_k \triangleq \{\mb{s_k}, \mb{s_{k-1}},...,\mb{s_0},\mb{a_k}, \mb{a_{k-1}},...,\mb{a_0}\}$.  That is, S-DTCBF is the traditional DTCBF condition in expectation. 
This constraint has been shown to provide bounds on the $K$-step exit probability \eqref{eq:k_step_safety} in a variety of contexts \cite{cosner2023robust,cosner2024freedman} and is often enforced on the system in the form of the following safety filter: 
\begin{align}
    \mb{u}^*_k = \argmin_{\mb{u} \in \mathcal{U}} & \quad \Vert \mb{u} - \mb{k}_\textup{nom}(t)\Vert\\
    \textup{s.t. } & \quad \E[~h(\mb{x}_{k+1}) ~|~ \mathscr{F}_k] \geq \alpha h(\mb{x}_k).   \nonumber
\end{align}
Unlike standard applications of CBFs, this optimization problem may be computationally complex and non-convex. 
Thus, modifications involving Jensen's inequality and generative modeling can be made to improve computational efficiency for hardware applications \cite{cosner2023robust, cosner2024generative}.

In this work, we specifically consider the probability bounds as generated by: 
\begin{theorem}[Freedman's Inequality for Stochastic Safety {\cite[Thm. 3]{cosner2024freedman}}] \label{thm:freedmans}
    If, for some $K \in \mathbb{N}_1$, $\sigma > 0$ and $\delta > 0$, the following bounds on the difference between the true and predictable update and the conditional variance hold for all $k \leq K$:
    \begin{align}
        \textup{Var}(h(\mb{x}_{k+1}) ~|~ \mathscr{F}_k)&\leq\sigma^2 \label{assp:bounded_var}\\
        \E[h(\mb{x}_k)~|~\mathscr{F}_{k-1}]-h(\mb{x}_k)&\leq\delta \label{assp:bounded_jumps} 
    \end{align}
    and the dynamics are constrained as in  \eqref{eq:dtcbf_ineq}
    for some $\alpha\in(0, 1)$, then the K-step exit probability is bounded as:
    \begin{align}
        P_u(K, \mb{x}_0)\leq e^{\frac{\alpha^K h(\mb{x}_0)}{\delta}}\left( \frac{\sigma^2 K}{\lambda}\right)^{\frac{\lambda}{\delta^2} } 
    \end{align}
    where $\lambda = \alpha^K h(\mb{x}_0) \delta + \sigma^2 K $.
\end{theorem}

To apply this theorem, we require two assumptions: first, a bound on the safety variance as in \eqref{assp:bounded_var}; second, a bound on the difference between the true safety value $h(\mb{x}_k) $ and the expected value as in \eqref{assp:bounded_jumps}. 
The first assumption is not very restrictive and allows for a large class of potential functions $h$, dynamics, and disturbance distributions. The second assumption is more restrictive, but generally applies in our setting, as the worst-case falling behavior would lead to a bounded difference between the commanded and true reduced-order-model behavior. 

\section{Disturbance Learning} \label{sec:learning}

While theoretical frameworks such as Freedman's inequality (Thm. \ref{thm:freedmans}) provide powerful methods for analyzing and synthesizing risk-aware controllers, their guarantees fundamentally depend on accurate characterization of the disturbance distribution $\mathcal{D}$. Rather than assuming that this distribution is known \textit{a priori} or constrained to a simplified parametric form (e.g., additive Gaussian noise), we propose a data-driven approach, based on \cite{cosner2024generative}, that leverages deep generative modeling to learn these distributions directly from empirical trajectories of the system. This approach enables us to capture complex, non-Gaussian, and state-dependent uncertainty patterns that more faithfully represent the actual disturbances encountered during hardware operation.

\newsec{Conditional Variational Inference.}  To account for the dynamics residual, we seek to train a generative model to approximate the dynamics residual distribution. To do this, we first collect a dataset of state, command, and disturbance tuples $\mathfrak{D} = \{ (\mb{x}_i, \mb{u}_i, \mb{d}_i) \}_{i=1}^N $.  We then train a Conditional Variation Autoencoder (CVAE) \cite{sohn2015learning} on this dataset with latent size 3 which yields a generative disturbance model $p_\theta(\mb{d}_k|\mb{x}_{k:k-N}, \mb{u}_{k: k-N})$. In contrast to previous work \cite{cosner2024generative}, the model is conditioned on a context window of length $N \in \mathbb{N}$, to allow the model to better capture temporal effects such as higher state derivatives or time delays. We find that providing this context greatly boosts modeling accuracy for a complex system like a humanoid robot (Sec. \ref{sec:results}). Note that the input $\mb{u}_i$ here is the unfiltered command, meaning we do not need to solve the algebraic loop of the filtered input \eqref{eq:optimal_tracking_theory}. 

We note that any class of generative disturbance model (e.g., diffusion \cite{ho2020denoising}, flow matching \cite{lipman2023flowmatchinggenerativemodeling}) can be used with our proposed safety framework (Sec. \ref{sec:safety}) - for SHIELD we choose to use CVAEs due to their expressivity and fast inference time, as shown empirically in \cite{cosner2024generative}.

\newsec{Stochastic Tracking with Learned Disturbance.} \label{subsec:tracking} SHIELD distinguishes itself from conventional safety layers through how it modulates control signals. While traditional approaches \cite{ames2014control, ames2019control,cosner2024generative} operate by modifying low-level signals (such as joint torques or raw actuation commands) to maintain safety, SHIELD instead modulates higher-level signals, i.e., the reference commands provided to the reduced-order model. 
This architecture is similar to that of a \emph{reference governor} \cite{garone2017reference}, which modulates reference or command signals into the controller/plant; the key difference is we modulate these signals with a CBF and without knowledge of the actual controller and plant dynamics.  
This modification enables the definition of safety constraints on simpler, more semantically meaningful states, making the system both more interpretable and manageable.

SHIELD recognizes that the ultimate objective is to achieve the intended system behavior, meaning the system accurately tracks the reduced-order model's trajectory.
To derive this ``best-tracking'' control, we define the optimal control as minimizing the expected difference between the next state of the reduced-order model under the desired command, and the next state of the actual system:
\begin{equation} \label{eq:optimal_tracking_theory}
    \mb{u}_k^* = \argmin_{\mb{u}_k \in \mathcal{U}} ~ \E [||\overline{\mb{x}}_{k+1} - (\mb{F}(\mb{x}_k) + \mb{G}(\mb{x}_k)\mb{u}_k + \mb{d}_k ) ||^2 | \mathscr{F}_{k}] \nonumber 
\end{equation}
where $\overline{\mb{x}}_{k+1}$ is the desired next position. 
Assuming pseudo-invertibility of $\mb{G}(\mb{x}_k)$, the optimal $\mb{u}$ is\footnote{The derivation of this follows from the equality $\E [||\overline{\mb{x}}_{k+1} - (\mb{F}(\mb{x}_k) + \mb{G}(\mb{x}_k)\mb{u}_k + \mb{d}_k ) ||^2 | \mathscr{F}_{k}] = ||\overline{\mb{x}}_{k+1} - (\mb{F}(\mb{x}_k) + \mb{G}(\mb{x}_k)\mb{u}_k + \E[\mb{d}_k|\mathscr{F}_k] ) ||^2 + \E[\Vert \mb{d}\Vert^2|\mathscr{F}_k] - \Vert\E[\mb{d}|\mathscr{F}_k]\Vert^2  $. Since this is true, it suffices to find the optimal $\mb{u}$ for $||\overline{\mb{x}}_{k+1} - (\mb{F}(\mb{x}_k) + \mb{G}(\mb{x}_k)\mb{u}_k + \E[\mb{d}_k|\mathscr{F}_k] ) ||^2 $ which is \eqref{eq:optimal_tracking_theory}.}:
\begin{align}
    \mb{u}_k^* = \mb{G}^\dagger(\mb{x}_k) ( - \mb{F}(\mb{x}_k) + \overline{\mb{x}}_{k+1} - \E[\mb{d}_k|\mathscr{F}_k]). 
\end{align}
However, since we do not have access to the true expectation $\E[\mb{d}_k|\mathscr{F}_k]$, we approximate this with the learned expectation computed from samples generated by the CVAE: 
\begin{align}
    \mb{u}_k^* = \mb{G}^\dagger(\mb{x}_k) ( - \mb{F}(\mb{x}_k) & + \overline{\mb{x}}_{k+1} \nonumber\\
    & - \E_{p_\theta}[\mb{d}_k|\mb{x}_{k:k-N}, \mb{u}_{k:k-N}]). 
    \nonumber
\end{align}
This $\mb{u}_k^*$ uses the learned disturbance distribution to select the command which reduces the mean squared error to the desired next state $\overline{\mb{x}}_k$. 

\begin{figure}[t]
    \vspace{1em}
     \centering
    \includegraphics[width=0.9\linewidth, center]{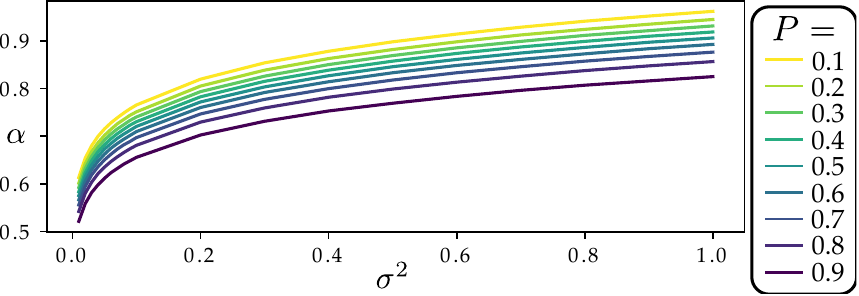}
    \caption{
    Higher $\alpha = L(P, K=10, h(x_0)=10, \delta=0.01, \sigma)$ values correspond to more conservative behavior. This increased conservatism is a  consequence of a lower $K$-step exit probability or a higher variance.
    }
    \label{fig:alphas}
      \vspace{-4em}
\end{figure}

\newsec{Safety with Learned Disturbance.}  In addition to using the learned dynamics residual to improve tracking, we can also use it to improve safety.
To do this, we select a maximum allowable risk level $P\in (0, 1)$. Given the horizon length $K \in \mathbb{N}$,  the initial safety value  $h(\mb{x}_0)$, the step-wise bound $\delta$ from assumption \eqref{assp:bounded_jumps}, and the variance bound $\sigma$ from assumption \eqref{assp:bounded_var} we can solve for the $\alpha$ that will result in the desired risk level bound $P$: 
\begin{align} \label{eq:alpha_picker}
    \alpha = L(P, K, h(x_0), \delta, \sigma)
\end{align}
In practice we approximate $L: (0,1) \times \mathbb{N} \times \R_{>0} \times \R_{>0} \times \R_{>0} \to (0, 1) $ numerically by performing root finding with Brent's Method \cite{brent2013algorithms} due to the complexity of the analytic solution. 
Evaluations of the function for different $\alpha$ values for a range of $P$ and $\sigma$ can be found in Fig. \ref{fig:alphas}. 

To apply Theorem \ref{thm:freedmans}, we address assumption \eqref{assp:bounded_var} and \eqref{assp:bounded_jumps} in turn, and how they apply to our application 
\begin{enumerate}
    \item For assumption \eqref{assp:bounded_var}, the variance bound $\sigma^2$ is approximated from the sampled dataset $\mathfrak{D}$,
    \item For assumption \eqref{assp:bounded_jumps}, we derive a bound from our application to bipedal robots. In this case, we can bound our difference between the true and predicted update for $h(\mb{x}_k)$ based upon the maximum step distance which can be measured in practice:
\begin{align} 
    \delta \triangleq 2(h(\mb{x}_{\textup{footstep}~k}) - h(\mb{x}_{\textup{footstep}~k+1})).
    \label{eq:delta_approx}
\end{align} 

\end{enumerate}

\begin{figure}
  \vspace{1em}
    \centering
    \includegraphics[width=\linewidth]{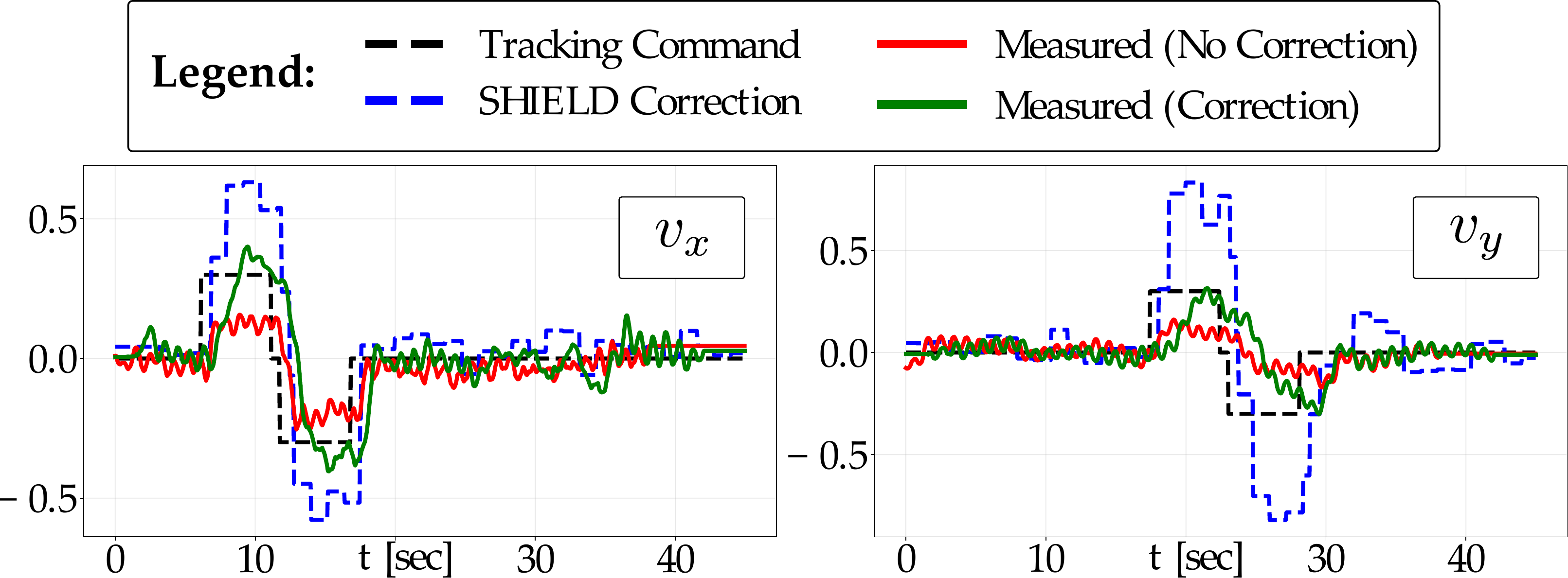}
    \caption{SHIELD improves tracking performance by correcting learned disturbances. After applying the SHIELD correction as shown by the blue dashed lines, the robot's tracking of the user's intended velocities (shown as a black dashed lines) improves.}
    \label{fig:tracking}
    \vspace{-15pt}
\end{figure}

In addition to meeting assumptions \eqref{assp:bounded_var} and \eqref{assp:bounded_jumps}, we must also enforce the \ref{eq:dtcbf_ineq} inequality, which we incorporate as a constraint in a safety filter of the form: 
\begin{align}
    \mb{u}_\textup{safe}^* = &\argmin_{\mb{u} \in \mathcal{U}}   \quad \Vert \mb{u} - \mb{u}^* \Vert\\
      &~\textup{s.t. }  \quad \E[h(\mb{F}(\mb{x}_k) + \mb{G}(\mb{x}_k) \mb{u}_k + \mb{d}_k) | \mathscr{F}_k ] \geq \alpha h(\mb{x}_k)   \nonumber
\end{align}

To enforce this constraint, we need to be able to quickly evaluate or lower bound the expectation $\E[h(\mb{F}(\mb{x}_k) + \mb{G}(\mb{x}_k) \mb{u}_k + \mb{d}_k) | \mathscr{F}_k ] $. To do this for concave $h$ functions, we employ Jensen's inequality as in \cite{cosner2023robust} to arrive at the following inequality:

\begin{proposition}[Probabilistic Invariance with Concave Safety Functions {\cite[Lem. 1]{cosner2023robust}}] \label{lem:rss}
    Consider a twice-continuously differentiable, concave function $h: \R^{n_x} \to \R$ with $\sup_{\mb{x} \in \R^{n_x}} \Vert \nabla^2 h(\mb{x}) \Vert_2 \leq \lambda_\textup{max} $ for some $\lambda_\textup{max} \in \R_{\geq 0 }$, and a random variable $\mb{x}$ that takes values in $\R^{n_x}$ with $\E[\Vert \mb{x} \Vert_2 ] < \infty $ and $\Vert \textup{cov}(\mb{x}) \Vert < \infty $. This function $h$ and random variable $\mb{x}$ satisfy: 
    \begin{align}
        \E[h(\mb{x}) ] \geq h(\E[\mb{x}]) - \frac{\lambda_\textup{max}}{2} \textup{tr}(\textup{cov}(\mb{x})). 
    \end{align}
\end{proposition}
This allows us to enforce the \eqref{eq:dtcbf_ineq} for concave, continuously differentiable $h$ indirectly by instead enforcing the tightened constraint: 
\begin{align}
    & h(\mb{F}(\mb{x}_k) + \mb{G}(\mb{x}_k) \mb{u}_k + \E_{p_\theta}[\mb{d}_k| \mb{x}_{k:0}, \mb{u}_{k:0} ])   \nonumber \\
    & \quad \quad \quad \quad - \frac{\lambda_\textup{max}}{2} \textup{tr}(\textup{cov}_{p_\theta}(\mb{d}_k |  \mb{x}_{k:0}, \mb{u}_{k:0}) \geq \alpha h(\mb{x}_k)   
\end{align}
where we can approximate $\E[\mb{d}_k|\mathscr{F}_k]$ and $\textup{cov}(\mb{d}_k|\mathscr{F}_k) $ using the learned dynamics residual distribution $p_\theta(\mb{d}_k | \mb{x}_{k:0}, \mb{u}_{k:0}) $. 

In summary, SHIELD uses the learned dynamics residual distribution from the CVAE to compute two distinct quantities: (1) an optimal input that minimizes the expected tracking error between the true system and desired next state, and (2) a minimal adjustment to this input that enforces probabilistic safety constraints. We emphasize that these components are fully modular. The tracking-optimized input can be used independently to reduce the sim-to-real gap, while the safety adjustment can be applied separately to enhance real-world safety guarantees. Alternatively, both components can be combined sequentially to simultaneously improve tracking performance and safety assurance, providing flexibility for different application requirements.
\section{Dynamic obstacle avoidance on stochastic reduced-order models}
\label{sec:safety}
In this section, we detail our approach to improve tracking and safety on a bipedal robot operating under random disturbances with a stochastic reinforcement learning-based controller $\bs{\pi}$.  
In particular, we use a PPO Actor-Critic learned controller $\bs{\pi}_\textup{PPO}$. This takes into account histories of proprioceptive and extereoceptive states $\mb{s}$ and a commanded velocity vector $\mb{u} = (v_x, v_y, \omega)$ using an LSTM and uses those to generate joint positions, $\mb{a}$.

To characterize the stochasticity of this controller, we use a CVAE to learn the distribution of the dynamics residual $\mb{d}$ conditioned on the last four\footnote{In practice, we condition on the last $N = \min(k, 4)$ states and commands for the algorithm to run at start time} system states and commands, i.e. ($\mb{x}_{k: k-3}, \mb{u}_{k: k-3}$). Specifically, we use a single integrator system with an additive disturbance as our simplified model:
\begin{align}
     \underbrace{
    \begin{bmatrix}
        p_x \\ p_y \\ \theta
    \end{bmatrix}_{k+1}}_{\mb{x}_{k+1}} = \underbrace{\begin{bmatrix}
        p_x \\ p_y \\ \theta
    \end{bmatrix}_{k}}_{\mb{F}(\mb{x}_k) } + \underbrace{
    \Delta_t \mb{I}_3}_{\mb{G}(\mb{x}_k) } \underbrace{\begin{bmatrix}
        v_x \\ v_y \\ \omega
    \end{bmatrix}_k}_{\mb{u}_k} + \underbrace{\Delta_t  \begin{bmatrix}
        d_x \\ d_y \\d_{\theta}
    \end{bmatrix}}_{\mb{d}_k}
\end{align}
where $p_x, p_y\in \R$, $\theta \in [0, 2 \pi)$, and $\Delta_t > 0 $ represent the $x$ and $y$ position, the yaw angle, and the state-update period and where $\mb{d}_k$ is a random disturbance that models the difference between the simplified model and the true dynamics. 

Using the Stochastic Tracking method detailed in Section \ref{subsec:tracking} leads us to the optimal tracking command:
\begin{equation}
    \mb{u}_{\text{adjusted}} = \
    \frac{\overline{\mb{x}}_{k+1} - \mb{x}_k }{\Delta_t}
    - \E_{p_\theta}[\mb{d}|\mb{x}_{k:k-3}, \mb{u}_{k:k-3}]
\end{equation}
where $\E_{p_\theta}[\mb{d}|\mb{x}_{k:k-3}, \mb{u}_{k:k-3}]$ is the mean disturbance learned by the CVAE. 
After modifying the command velocity with the predicted dynamics residual to improve tracking, we apply our safety filter which minimally modifies that command to enforce our safety constraint. For application, we consider obstacle avoidance with respect to $N \in \mathbb{N}$ obstacles as characterized by the signed distance function (sdf):
\begin{align}
    \textup{sdf}(\mb{x}) = \min_{i \in \{1, \dots N \} } \left \Vert 
        \begin{bmatrix}
            p_x\\
            p_y
        \end{bmatrix} 
        - \bs{\rho}_{i} \right \Vert - R_i
\end{align}
where $\bs{\rho}_i \in \R^2$ is the planar position of obstacle $i$ and $R_{i}>0$ is the robot radius plus the obstacle radius. 

To incorporate additional obstacles and reduce chattering oscillation that can occur when the closest obstacle switches, we smooth the SDF collision constraint to be: 
\begin{align} \label{eq:h_smooth}
    h_{\textup{smooth}}(\mb{x}_k) = \lambda(1-e^{-\gamma \textup{sdf}(\mb{x}_k)})
\end{align}
where $\lambda > 0$, $\gamma >0$ are positive constants controlling the maximum magnitude and smoothness of safety. 

Since we are only considering the closest obstacle, we make the following concave approximation: 
\begin{align}
    \widehat{h}(\mb{x}) = \lambda \left(1-e^{-\gamma( (\mb{p} - \bs{\rho}_{i})^T\mb{e}_i- R_i)} \right) \label{eq:convexification}
\end{align} 
\begin{align} 
\label{eq:h_tilde}
    \tilde{h}(\mb{x}) =
\begin{cases} 
    \widehat{h}(\mb{x}), ~~~~~~~~~~~\text{if } (\mb{p}-\bs{\rho}_{i})^\top \mb{e}_i\geq 0 \\
    \nabla_\mb{x}\widehat{h}(\mb{x}) + \lambda(1-e^{\gamma R_i}), ~~\textup{else}
\end{cases} 
\end{align}
where $\mb{p} \triangleq [p_x, p_y]^T$ and $\mb{e}_i \in \R^2$ with $||\mb{e}_i||_2 = 1$ is the unit direction towards the closest obstacle from the previous timestep, i.e. $(\mb{p}_k - \bs{\rho}_i ) / \Vert \mb{p}_k - \bs{\rho}_i \Vert$.   

In the case of a single obstacle, we provide the following bound which will allow us to build conditions that enforce a bound on the $K-$step failure probability in practice:
\begin{theorem}[Single-Obstacle Avoidance with Concave Barrier Functions]\label{thm:ccv_barrier_directional}
    Consider the function 
    $\tilde{h}$ as in \eqref{eq:h_tilde} with $N=1$ 
    and a random variable $\mb{x}$ that takes values in $\R^{n_x}$ with $\E[\Vert \mb{x} \Vert_2 ] < \infty $ and $\Vert \textup{cov}(\mb{x}) \Vert < \infty $. This function $\tilde{h}$ and random variable $\mb{x}$ satisfy: 
    \begin{align}
        \E \left[\tilde{h}(\mb{x}) \right] \geq \tilde{h}(\E[\mb{x}]) - \frac{\lambda_\textup{max}}{2}\mb{e}_1^T\textup{cov}(\mb{x})\mb{e}_1. 
    \end{align}

\end{theorem}
\noindent Please see the appendix for the proof.

\begin{algorithm}[t]
    \caption{SHIELD: Deployment Phase}
    \begin{algorithmic}[1]
    \State Initialize $k \gets 0, \mb{x} \gets \mb{x}_0$
    \State Initialize $P, \delta, \alpha$
    \While{true}
        \State $\textup{obstacles} \gets \{\bs{\rho}_1, ..., \bs{\rho}_M\}$
        
        \State $h_k \gets \min_i\tilde{h}(\mb{x}, \bs{\rho}_i)$, $i^* \gets \arg\min_i\tilde{h}(\mb{x}, \bs{\rho}_i)$
        
        \If{$k \textup{ modulo } K = 0$} 
            \State $\Sigma \gets \textup{cov}_{p_\theta}(\mb{d} | \mb{x}_{k:k-N}, \mb{u}_{k:k-N})$
            \State $\alpha \gets L(K, h_k, P, \delta, \Sigma)$
        \EndIf
        \State Get $\mb{u}_{\text{cmd}}$ as input
        \State $\mb{u}_{\text{adjusted}} \gets \mb{u}_{\text{cmd}} - \E_{p_\theta}[\mb{d}|\mb{x}_{k:k-3}, \mb{u}_{k:k-3}]$
        \State $\mb{e} \gets \frac{\mb{p}_k - \bs{\rho}_{\textup{obs}, i^*}}{||\mb{p}_k - \bs{\rho}_{\textup{obs},i^*}||}$, $\lambda \gets \lambda_{\max}(\bs{\rho}, \mb{e})$
        \State $\mb{u}_\textup{safe}^* \gets  \min_\mb{u} \Vert\mb{u} - \mb{u}_{\text{adjusted}}\Vert^2$ 
        \State \hspace{1cm} s.t. $\tilde{h}(\mb{F(x)} + \mb{G(x)u}) -\frac{\lambda}{2}\mb{e}^T\Sigma \mb{e}\geq \alpha h_k$
        \State Apply command $\mb{u}_\textup{safe}^*$, $\mb{x_k} \gets \mb{x_{k+1}}$, $k \gets k + 1$
    \EndWhile
    \end{algorithmic}
    \label{alg:main}
\end{algorithm}

This allows us to enforce the \eqref{eq:dtcbf_ineq} for concave, continuously differentiable $h$ indirectly by instead enforcing the tightened constraint: 
\begin{align}\label{ccv_barrier_constraint}
    & h(\mb{F}(\mb{x}_k) + \mb{G}(\mb{x}_k) \mb{u}_k + \E_{p_\theta}[\mb{d}_k| \mb{x}_{k:k-3}, \mb{u}_{k:k-3} ])   \\
    & \quad \quad \quad \quad - \frac{\lambda_\textup{max}}{2} \mb{e}_i^T\textup{cov}_{p_\theta}(\mb{d}_k |  \mb{x}_{k:k-3}, \mb{u}_{k:k-3})\mb{e}_i \geq \alpha h(\mb{x}_k)  \nonumber 
\end{align}
where we can approximate $\E[\mb{d}_k|\mathscr{F}_k]$ and $\textup{cov}(\mb{d}_k|\mathscr{F}_k) $ using the learned dynamics residual distribution $p_\theta(\mb{d}_k | \mb{x}_{k:k-3}, \mb{u}_{k:k-3}) $. In practice, we find that the utility of SHIELD generalizes to multiple obstacles; however, we leave a rigorous theoretical analysis of the nonconcave $\tilde{h}$ with multiple obstacles for future work.

\begin{figure}[t]
    \centering
    \includegraphics[width=\linewidth]{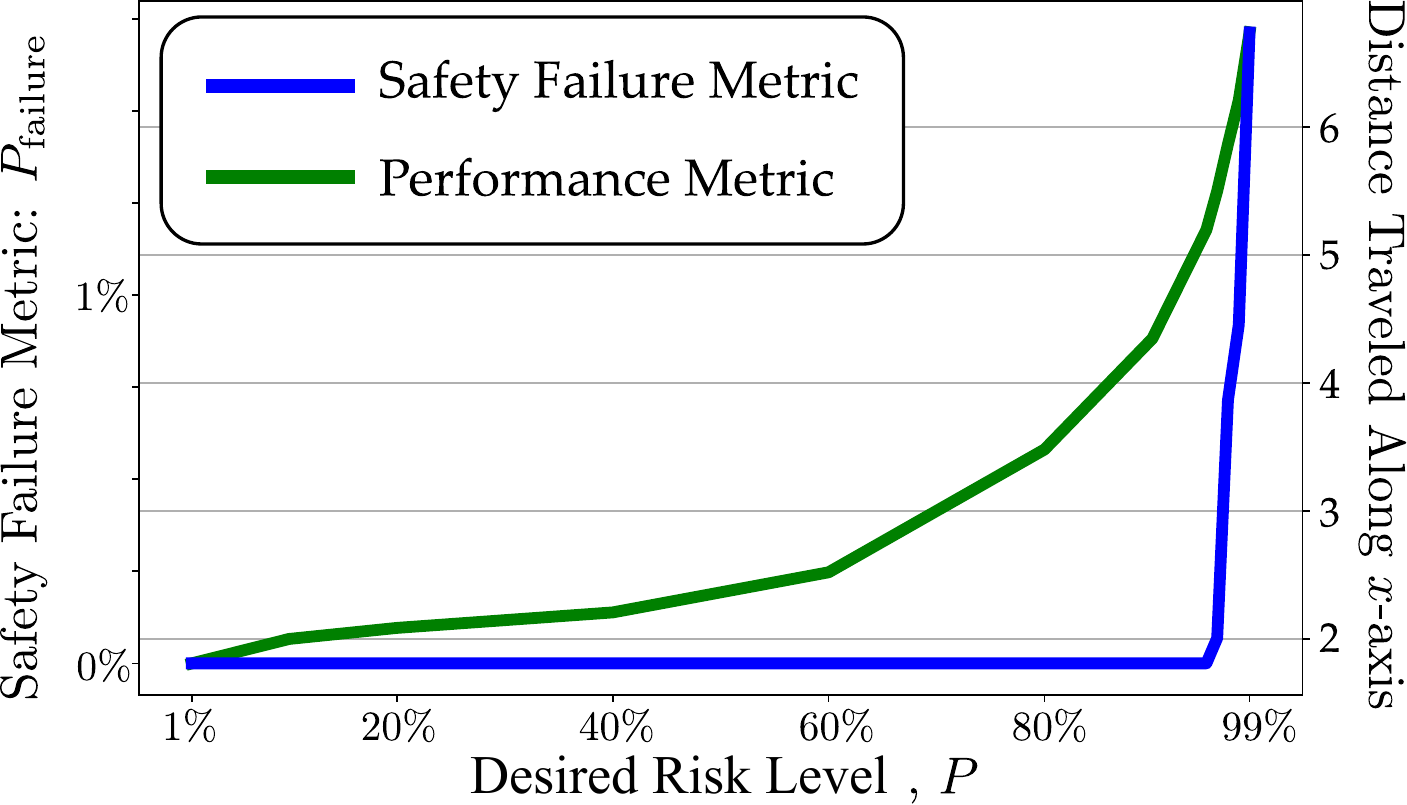}
    \caption{The trade-off between performance and safety.  As the probability of K-step exit increases, we achieve better performance at cost of an increasing amount of safety violation under the proposed metric.}
    \label{fig:sim_example}
    \vspace{-20pt}
\end{figure}

To determine the appropriate $\alpha $ to get a desired risk level across $K$ steps, we use the $L$ function in \eqref{eq:alpha_picker}. To calculate $\alpha$, a desired risk level $P$ is chosen, the current safety value is noted as $h(\mb{x}_k)$, the worst case $\delta$ is approximated as in \eqref{eq:delta_approx}, and the covariance $\sigma$ is set to the maximum value experienced in the experimental data.  Furthermore, to extend the guarantee beyond $K$ steps, we recalculate $\alpha$ every $K$ steps. Thus, each successive $K$ steps satisfies the bound in Prop. \ref{lem:rss} and they can be connected using the union bound: 
\begin{align} \nonumber
    \scalebox{0.84}{$\mathbb{P}\left \{ \min_{k\in[0, K\times F ]} h(\mb{x}_k) < 0 \right \}  \leq \sum_{i=0}^{F}\mathbb{P}\left\{\min_{k\in [Ki, K(i+1)]} h(\mb{x}_{k})<0\right\}$}
\end{align}
where $F$ is the number of $K$-step intervals in the experiment.
We show the SHIELD deployment stage in Alg. \ref{alg:main}.

\begin{figure*}[t]
    \includegraphics[width=\linewidth]{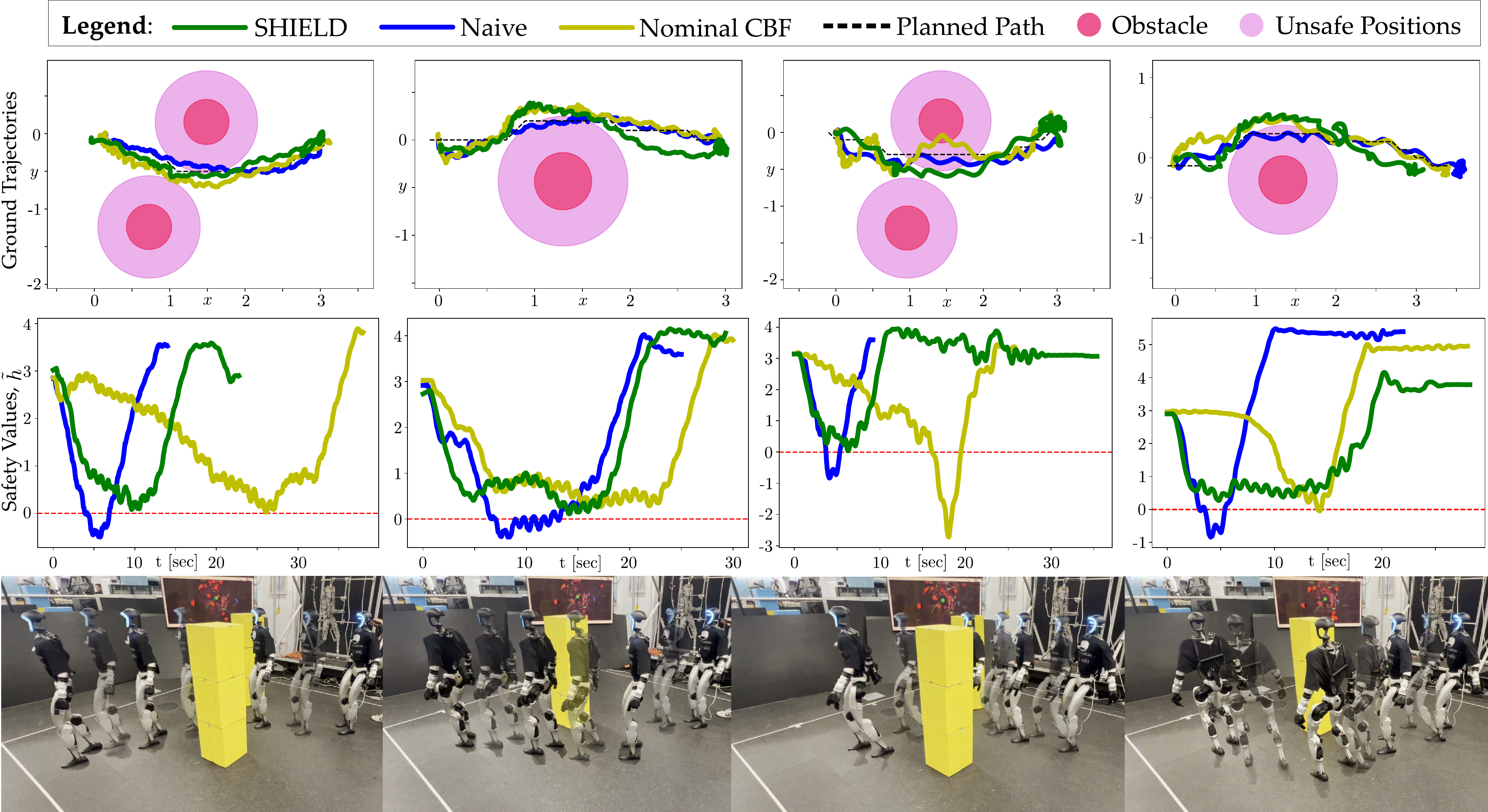}    
    \caption{SHIELD enforces safety in collision avoidance with adaptive conservatism. The A* planner path is not necessarily safe even though it does not cross the obstacle, thus naively following the path would result in collisions or scrapes. Nominal CBF, due to not accounting for the inaccurate reduced order model, would also result in collisions or be extremely conservative.}
    \label{fig:experiments}
    \vspace{-2em}
\end{figure*}

\vspace{-5pt}
\section{Experiments}
\label{sec:results}
We demonstrate the validity of SHIELD on a simple simulated system and then on a Unitree G1 humanoid robot, aiming to show the method's adaptable conservativeness, performance, and robustness.

\newsec{Simulation.} We first present a simplified simulation problem consisting of a single integrator disturbed by a 0-mean, multivariate student's $t$-distribution with clipped tails.
We randomize the placement of obstacles and the radii of the obstacles and the robot. 
The robot moves along the ${x}$-direction with a constant commanded velocity of 0.5m/sec, which we filter using SHIELD \eqref{alg:main}.

For all experiments, we set the discrete time difference $\Delta t = 0.01$ and the scaling factors in the safety function to be $\lambda = 10$, $\gamma = 0.5$, $K = 10$, and the upper bound on the assumption of bounded difference of \eqref{assp:bounded_jumps} to be 1.

We then simulate the system over $N_\textup{trials} = 100$  trials of $N_\textup{steps}= 2000 $ steps each and calculate the percentage of violations $P_\textup{failure}$ as:
\begin{equation}
P_\textup{failure} = \frac{\sum_{i=1}^{N_\textup{trials}}\sum_{j=1}^{N_{\textup{steps}}} \mathds{1}_{\{h(\mb{x}_{i, j})<0\}}}{N_{\textup{trials}} N_\textup{steps}}
\end{equation}
and quantify performance as total distance traveled in the commanded direction. 
We observe that the probability of $K$-step failure is a tuning knob to encourage more risky behavior at the cost of higher chance of collisions (though still well below the target percentage); see Fig. \ref{fig:sim_example}. 

\newsec{Hardware Setup.} The Unitree G1 humanoid robot has a height of 1.32 meters and weighs approximately 40kg, with 23 actuated degrees of freedom. We employ an onboard Jetson Orin NX for computation, a Livox Mid-360 LiDAR for sensing the environment, and an Intel T265 to localize the robot. Euclidean clustering \cite{Rusu_ICRA2011_PCL} is applied to the LiDAR pointcloud to locate obstacles of interest in the scene.

To test the generalization of SHIELD in deployment, we conduct experiments with two different walking controllers:
\begin{enumerate}
    \item \textbf{built-in}: the Unitree built-in controller \cite{unitree_sdk2}
    \item \textbf{custom}: We train a custom RL locomotion controller in IsaacLab \cite{mittal2023orbit} using standard rewards from \cite{gu2024advancing}.
\end{enumerate}

Approximately 6 minutes of training data are collected for each controller to train the CVAE for both the \textit{built-in} and \textit{custom} controllers. We query the CVAE to update the mean and covariance of the disturbance distribution at 0.83Hz, and we filter the command velocity at 100Hz. 

\newsec{Learned Tracking.} We first test the velocity tracking capabilities of the SHIELD framework. In these experiments, we send a pre-set sequence of velocity commands through the framework to the controller and compare our resulting velocities to the command sequence. We achieve noticeable improvements in tracking as shown in Fig. \ref{fig:tracking}.

\newsec{Obstacle Avoidance.} First, we conduct controlled experiments with fixed obstacles. 
We define success as the robot walking past obstacles without making contact. 
We model the detected obstacles as cylinders of radius 0.3m and the robot to have a safety margin of 0.38m from the center of mass. To navigate, we first use A* \cite{Hart1968} to first plan a path through free space, we then generate nominal velocities by directing the robot from its current position to the next node on the path and filter the commanded velocities with SHIELD. 
We present both single-obstacle and multi-obstacle cases. 
In single-obstacle experiments, naively following the A* path alone does not completely avoid obstacles due to state tracking errors.
The nominal DTCBF filter, being unaware of the dynamics residual, either performs too aggressively colliding with the obstacle, or performs to conservatively taking almost twice the time to arrive at the goal. However, SHIELD enables the robot to completely bypass the obstacle.
We observe similar behavior in multi-obstacle scenarios, where SHIELD is able to adjust conservativeness online to only enforce maximum safety conditions when needed, resulting in more dynamic behavior.
The results of these experiments can be seen in Fig. \ref{fig:experiments}. 

\newsec{Unstructured Outdoor Environment.} We also perform experiments in unstructured outdoor environments for further validation.
In these tests, a user provides joystick inputs to the robot for safety reasons and would either control the robot to walk directly towards people or provide no input and let the robot stay in place unless people encroach on its safety boundary. 
These experiments can be seen in Fig. \ref{fig:hero} and Fig. \ref{fig:pedstrian} and the experimental video \cite{video}. 

\section{Conclusion} 
\label{sec:conclusion}
This paper presented SHIELD: a safety layer that leverages stochastic discrete-time control barrier functions (S-DTCBF) to guarantee safety in probability.  Importantly, this can be added to an existing autonomy stack, wherein the dynamics of a nominal controller can be learned as the residual on a simplified model. SHIELD then filters the nominal commands to produce safe inputs as they are sent to the system via an S-DTCBF.  This framework is instantiated on a humanoid robot in the context of collision avoidance, where it is shown to outperform a nominal safety filter in hardware experiments on the Unitree G1 humanoid.  
This paper, therefore, demonstrates that by combining a general-purpose RL locomotion controller with a robot-specific stochastic safety layer, SHIELD achieves both high-performance walking and robust safety constraint satisfaction under uncertainty on humanoid robots.


\appendix

\subsection{Proof of Theorem \ref{thm:ccv_barrier_directional}} \label{apdx:proof}
\begin{proof}
    Consider the convex, twice differentiable function $\eta:\R^{n_x}\rightarrow\R$ defined as $\eta = -\tilde{h}$. By second-order Taylor's theorem, for $\mb{x}, \bs{\mu} \in \R^n$ there exists an $\omega \in (0, 1)$ such that:
    \begin{align}
        \eta(\mb{x}) = \eta(\bs{\mu}) + \nabla\eta(\bs{\mu})^T\mb{d} + \frac{1}{2}\mb{d}^T\nabla^2\eta(\mb{c})\mb{d}
    \end{align}
    where $\mb{d} = \mb{x}-\bs{\mu}$ and $\mb{c} =\omega \mb{x} + (1-\omega) \bs{\mu}$. From the construction of $\tilde{h}$, the Hessian of $\eta $ is : 
    \begin{align} \label{eq:hessian}
        \nabla^2 \tilde{h}(\mb{x}) = \begin{cases}
             \varphi(\mb{x}) \mb{e}_1 \mb{e}_1^\top, & \textup{ if } (\mb{p} - \bs{\rho}_1)^\top \mb{e}_1 \geq 0 \\ 
            0,  & \textup{else}.
        \end{cases}
    \end{align} \label{lambda_max}
    where $\varphi(\mb{x}) \triangleq \gamma^2 \lambda e^{- \gamma ((\mb{p} - \bs{\rho}_1)^\top \mb{e}_1 - R_1) }$ which is bounded over the if case $\{~\mb{x} \in \R^{n_x} ~|~ (\mb{p} - \bs{\rho}_1)^\top \mb{e}_1 \geq 0 ~\}$, we call this bound $\lambda_\textup{max}> 0 $.

    Here $\nabla^2\tilde{h}(\mb{x}) $ is a diagonalizable, positive semi-definite matrix and when it has a non-zero eigenvalue, the associated eigenvector is equal to $\mb{e}_1$. Therefore: 
    \begin{align}
        \eta(\mb{x}) 
        & \leq  \eta(\bs{\mu}) + \nabla\eta(\bs{\mu})^\top\mb{d} + \frac{\lambda_\textup{max}}{2}\mb{d}^\top\mb{e}_1\mb{e}_1^\top \mb{d}
    \end{align}

    Next, we follow the proof of \cite[Lem. 1]{cosner2023robust} with $\bs{\mu} = \E[\mb{x}] $: 
    \begin{align}
        \E& [\eta(\mb{x}) ] - \eta(\E[\mb{x}]) = \int_{\R^{n_x}} (\eta(\mb{x})- \eta(\bs{\mu}) )p(\mb{x}) d\mb{x}\\
        & \leq \int_{\R^{n_x}} \nabla(\bs{\mu})^\top \mb{d} + \frac{\lambda_\textup{max}}{2} \textup{tr}(\mb{e}_1^\top\mb{d}\mb{d}^\top \mb{e}_1) p(\mb{x}) d\mb{x}\\
        &= \frac{\lambda_\textup{max}}{2} \textup{tr}(\mb{e}_1^\top \textup{cov}(\mb{x})  \mb{e}_1) = \frac{\lambda_\textup{max}}{2}\mb{e}_1^\top \textup{cov}(\mb{x}) \mb{e}_1.
    \end{align}
\end{proof}

\balance
\bibliographystyle{IEEEtran}
\bibliography{cosner}

\end{document}